# A simple and efficient SNN and its performance & robustness evaluation method to enable hardware implementation

Anmol Biswas[°], Sidharth Prasad[+], Sandip Lashkare, and Udayan Ganguly

Department of Electrical Engineering
Indian Institute of Technology, Bombay
Mumbai, India
anmolbiswas@gmail.com, sidharthprasad@iitb.ac.in, lashkaresandip@gmail.com, udayan@ee.iitb.ac.in

*Abstract—* **Spiking Neural Networks (SNN) are more closely related to brain-like computation and inspire hardware implementation. This is enabled by small networks that give high performance on standard classification problems. In literature, typical SNNs are deep and complex in terms of network structure, weight update rules and learning algorithms. This makes it difficult to translate them into hardware. In this paper, we first develop a simple 2-layered network in software which compares with the state of the art on four different standard data-sets within SNNs and has improved efficiency. For example, it uses lower number of neurons (3 ×), synapses (3.5×) and epochs for training (30×) for the *Fisher Iris* classification problem. The efficient network is based on effective population coding and synapse-neuron co-design. Second, we develop a computationally efficient (15000×) and accurate (correlation of 0.98) method to evaluate the performance of the network without standard recognition tests. Third, we show that the method produces a robustness metric that can be used to evaluate noise tolerance.**

## I. INTRODUCTION

Artificial neural network (ANN) algorithms (e.g. deep learning) are capable of recognizing complex patterns (e.g. voice, image etc.) but run on von-Neumann hardware in servers. Hence, the high power (~MWatt) requirements of the servers pale in comparison with that of the brain (~100 Watts) [1]. Hence a dedicated hardware implementation inspired by the brain architecture is highly desirable. Spiking Neural Networks (SNN) are being extensively researched to solve classification and machine-learning problems and produce neuromorphic hardware. Hardware implementable algorithms need three features. *Firstly, some learning rules may be mathematically complex for simple hardware implementation. Hence simpler alternatives are selected.* For example, some spiking neural networks make use of back-propagation type algorithms for supervised learning and classification ([2] [3] [4] [5] [6]). This requires a centralized computation of local weight changes for global error reduction by back-propagation for learning – which leads to inefficiencies. In addition, artificial algorithmic constraints on the behavior of the neurons, like restricting them to generate limited number of spikes (like only one or two) in the training interval, and complex networks by splitting each synapse into multiple synapses with different time delays [2] [3] [4] increase the complexity of implementation. In comparison, a biomimetic, simple and local Spike Time Dependent Plasticity (STDP) - like Supervised Hebbian Learning [7] [8] [9] is used for local weight change computation in our work. Thus, STDP based networks can be translated into hardware more efficiently. *Secondly, certain algorithms cannot be simply translated to a cross-bar architecture.* For example, the ReSuMe approach [8] is based on STDP rules, but requires a special class of neurons called "teacher neurons". This requires *3 inputs to every synapse* (pre-neuron, post-neuron & teacher neuron) which is difficult to implement in a cross-point architecture where 2 inputs (a row and column driver) normally serve every synapse. *Thirdly, the networks must be small as hardware level simulations are computationally expensive*. For example, various networks [4] used for Fisher Iris data set use ~100 neurons and ~500 synapses. The computational time for training our two-layered network using transient simulation is a circuit simulator – SPICE, is roughly 2 hours, which is large, despite the simplicity [10]*. After training, performance evaluation requires more computation. Thus, to avoid complexity and computational cost for hardware simulation, a simple and small network with high performance on *standard classification problems* is needed. In addition, a more efficient strategy to evaluate the performance of the SNN is attractive.

In this paper, we firstly develop a small and efficient network which matches state-of-the art performance on *two* classification problems and performs well on another two and identify the related design principles. Second, we develop a highly efficient method of evaluating the network performance after learning. Finally, we evaluate the noise tolerance by stochastic simulations enabled by efficient performance evaluation.

## II. APPROACH

*A. Network*

Our proposed network is small and simple. It consists of only two layers - one input layer and one output layer, with no hidden layers, as illustrated in Fig. 1. In a classification problem

---

[+] [°] First authors with equal contributions

(say *malignant* vs. *benign* cancer), a database (say Wisconsin Cancer dataset) has $S$ input features (e.g. patient heat-rate or temperature etc.), where each feature has a range of values (say heartrate is 70-100 pulses/min). First, a population coding scheme ([11] [12] [13]) consisting of normalization and transformation steps is used to input the data as corresponding current levels into the input layer as described later. The input layer is fully connected to the output layer through excitatory synapses i.e. each input neuron is connected to all output neurons and vice versa. The output neurons are interconnected among themselves in pairs through inhibitory lateral synapses to implement the *Winner takes all* rule [14]. Thus, only one neuron spikes which indicates that the network has identified the input to be of the same class as associated with that output neuron. During learning, in the training phase, negative bias currents to the output layer are used for supervision to discourage the neuron associated with the wrong class from firing.

We consider two types of feature transformations – (i) Gaussian transformations and (ii) linear transformations for mapping the normalized feature values to the constant current values that is input to input neurons. $F$ input features undergoing $T$ transformations will require $F \times T$ input neurons.

*a) Gaussian transformations*

The interval $[0,1]$ of $x$ (i.e. the feature value after normalization) is divided into $T - 1$ sections, with a Gaussian Radial Basis Functions (RBF) centered at each section boundary. Thus, we have a total of $T$ Gaussians that generate the current to be fed to the input neurons. The RBF is given by:

$$I_j(x) = I_{max} \cdot exp\left(\frac{x - \mu_j}{2\sigma^2}\right) \qquad (2)$$

where $I_{max}$ is a maximum input current value. The methodology to select $I_{max}$ is discussed later. Each of these RBFs act as level-sensitive sensors, as they pick up the signal only if it lies in a certain range where its response in non-zero, as illustrated in Fig. 2

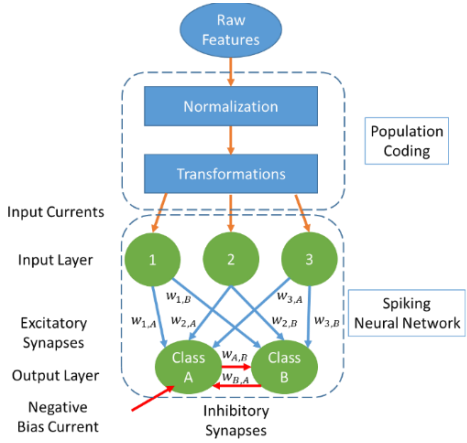

Fig. 1. The network structure with two layers. The raw features are first normalized between 0 and 1 and then undergo transformations, before being input to the first layer. The first layer spikes propagate to the output layer via the excitatory synapses. Negative bias current is used for supervision while lateral inhibitory synapses implement the 'Winner takes all' rule. Each output neuron corresponds to a particular sample class. $w_{i,j}$ is the weight of the respective synapse.

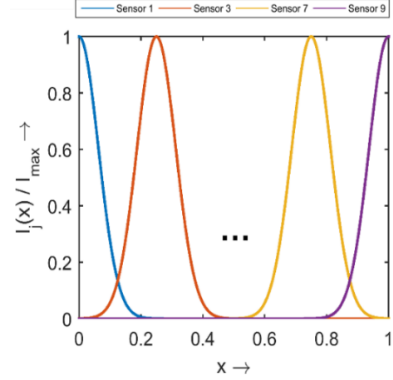

Fig. 2. The Gaussian Transformations with a total of 9 sensors (The other transformations are not shown for clarity). Each transformation produces a large output if the input lies in a certain range and thus acts as a range-specific sensor.

*b) Linear function transformations*

1. *High value sensor* causes spiking when the value of the feature is high.
$$I_1(x) = I_{max} \cdot x \qquad (3)$$

2. *Low value sensor* causes spiking when the value of the feature is low.
$$I_2(x) = I_{max} \cdot (1 - x) \qquad (4)$$

3. *Intermediate Value sensor* causes spiking when the value of the feature lies close to the middle.
$$I_3(x) = I_{max} \cdot (1 - 2 \cdot |x - 0.5|) \qquad (5)$$

4. *Extreme Value sensor* causes spiking when the value of the feature lies away from the middle.
$$I_4(x) = I_{max} \cdot 2 \cdot |x - 0.5| \qquad (6)$$

*B. Normalization*

The raw feature values are mapped on a scale of 0 to 1, given by the following equation.

$$x_i = \frac{z_i - \min(z)}{\max(z) - \min(z)} \qquad (1)$$

where $z$ is the raw feature value and $x$ is the normalized feature value.

*C. Transformations*

To implement learning in a two-layer network with current based inputs, we had to consider ways to make the input neurons respond to mid and low level/intensity values of the samples' features. Directly encoding the intensity of the feature values with input currents is insufficient as it makes the network blind to mid-level and low-level feature values.

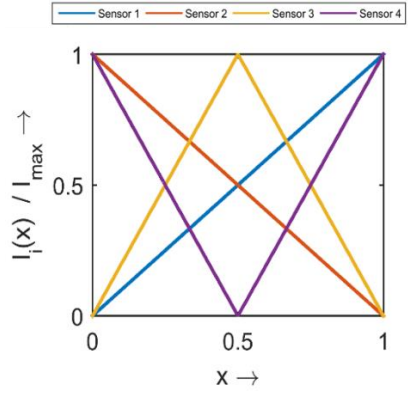

Fig. 3. The four linear transformations: a) High value sensor b) Low value sensor c) Intermediate value sensor d) Extreme value sensor

### D. Neurons

We use the Leaky Integrate and Fire (LIF) model to model the neurons in the network [15]. It is a simple Resistor-Capacitor circuit that fires a spike when the potential reaches the threshold value ($V_t$) and immediately after spiking, the potential is reset to the Resting Potential ($E_l$). The rate of firing is thus, a function of the applied current. Eq. 7 describes the dynamics of the LIF neuron.

$$\frac{dV}{dt} = \frac{1}{C}(-g(V - E_l) + I) \quad (7)$$

where $V$ is the membrane potential, $C$ is the capacitance, $g$ is the conductance, $E_l$ is the resting potential and $I$ is the input current to the neuron.
Values used: $= 300pF$, $E_l = -70mV$, $g = 30nS$ and $V_t = 20mV$.

### E. Synapses

A synapse connects a pair of neurons i.e. input to output neuron. It converts a voltage spike from pre-neuron to a current response in the post-neuron whenever the pre-neuron fires a spike. The form of the current pulse is given by the following equation.

$$I(t) = w \cdot I_0 \cdot \left(\exp\left(-\frac{t-t_s}{\tau_m}\right) - \exp\left(-\frac{t-t_s}{\tau_s}\right)\right) \quad (8)$$

where, $w$ is the weight of the synapse, $t_s$ is the time instant when the pre-neuron fires a spike, $\tau_s$ and $\tau_m$ are time constants with $\tau_m > \tau_s$ and $I_0$ is a constant current to scale the synaptic current pulse to the required levels for the LIF neurons [16].

We use two kinds of synapses in the network- (i) excitatory and (ii) inhibitory. They both have similar behavior, except that the weights associated with the excitatory synapses are positive (i.e. $w > 0$, they promote post-neuron spiking by pushing current towards it) while those associated with the inhibitory synapses are negative (i.e. $w < 0$, they inhibit post-neuron spiking by pulling out current from it).

Values used: $I_0 = 10pA$, $\tau_m = 10ms$ and $\tau_s = 2.5ms$ for excitatory synapses. $I_0 = 0.1nA$, $\tau_m = 50ms$.

### F. Learning Rule

The time difference between the pre-neuron spike and the post-neuron spike, $\Delta t$ is defined by $(t_{post} - t_{pre})$ as shown in Fig. 4 (a). In STDP, the weight change ($\Delta w$) is a function of $\Delta t$ between pre- vs. post-neuron spike as shown in Fig. 4 (b). However, in realistic RRAM device (Fig. 4 (c)), RRAM conductance (equivalent to weight, $w$) first increases linearly with increasing number of pulses (akin to ideal STDP) and then saturates. Hence, we use a learning rule (Eq. 3-4) used which falls under the category of Supervised Hebbian Learning (SHL) [7]. Here, weight saturates as it gets closer to their maximum or minimum values akin to realistic RRAM devices.

$$\Delta w = A_{up} \cdot \left(1 - \frac{w}{w_{max}}\right)^\mu \cdot \exp\left(\frac{-\Delta t}{\tau_{up}}\right), \Delta t > 0 \quad (9)$$

$$\Delta w = A_{down} \cdot \left(\frac{w}{w_{max}}\right)^\mu \cdot \exp\left(\frac{\Delta t}{\tau_{down}}\right), \Delta t < 0 \quad (10)$$

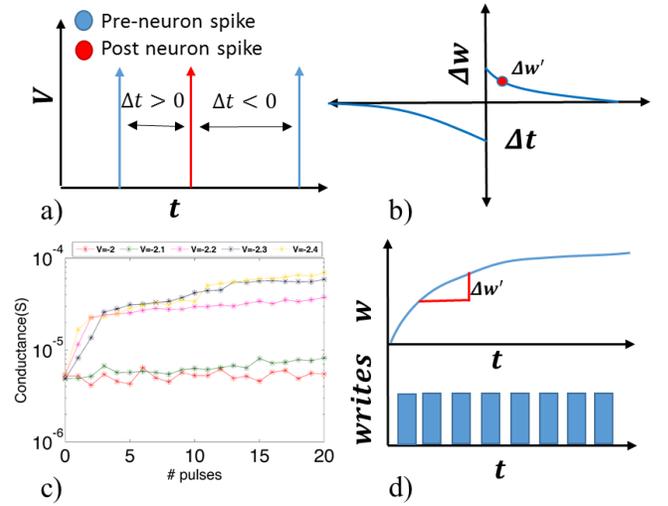

Fig. 4. STDP rule a) Showing the two possible cases, causal and anti-causal spiking b) Standard STDP rule c) Experimental plot of **Resistance** vs **no. pulses** showing weight saturation d) weights saturation in our model as multiple writes are performed showing the same observation as in (c)

Also, $A_{up}$ is positive (for potentiation, there should be an increase in weight) while $A_{down}$ in negative (for depression, there should be a decrease in weight) and μ should be positive and greater than 1 to enable control over the rate of saturation. [17]
Values used: Different values of $A_{up}$, $A_{down}$ (in the range: 1 to 10 and -1 to -20 respectively) and $w_{max}$ were experimented with (values of $w_{max}$ were kept between 300 to 1000 for good results), $\tau_{up} = 10ms$ and $\tau_{down} = 20ms$ were used.

### G. Weight range and Initialization

The parameter $w_{max}$ is the value of maximum weight that the excitatory synapses could take while the minimum value is zero. We found that, when using *more* transformations, the optimum range of $w_{max}$ values is lower than when using fewer transformations. Alternatively, we can also increase the Threshold Potential in case of larger number of input neurons.

Generally speaking, we found that keeping the initial mean value between one-third and half of $w_{max}$ gave good results. The third parameter is the initial weight distribution i.e. mean of the initially assigned random weights. We found that the quality of learning (measured by the classification accuracy) depended heavily on the initial conditions. This happened because, for the network to learn, it should be able to spike and so if the initial weights are too low, the network cannot learn at all

## H. Training and testing

A small subset of all the samples was taken for training the network (details for each dataset is presented later). Each of these training samples was shown repeatedly to the network for a fixed duration of 100 $ms$. All training samples are serially shown to the network in an *epoch*.

To test performance after learning, the bias currents are set to zero and the entire dataset is shown to the network. Within the testing time interval allotted to each sample, the recognition is counted as successful only when the correct neuron (corresponding to the class to which the sample actually belongs) fires and the others do *NOT* fire [14].

## III. PERFORMANCE AND MECHANISM

### A. Benchmarking Performance of different datasets

We extensively tested our SNN algorithm on *four* standard classification problems – the *Fisher Iris* [18], *Wisconsin Breast Cancer* [19], *wine dataset* [20] and *heart-statlog dataset* [21]. For each problem, we evaluate the effect of different transformations in the networks on their performance. We also benchmark our algorithm with other neural network based methods from literature, in terms of network size, number of training iterations required and the classification accuracy achieved. Note that the classification accuracy is stated as a 10 experiment average to account for stochasticity in selection of training samples.

#### a) Fisher Iris classification dataset

The *Fisher Iris* dataset consists of 150 samples and each sample possesses 4 features. The samples belong to three different classes, 50 to each class. Two of these classes are *not* linearly separable [18]. 15 randomly chosen samples from each class, i.e. a total of 45 samples were taken for training and the learned weights are tested on the entire dataset of 150 samples.

TABLE I. summarizes the results. We observe that the network with 4 linear sensors achieves the best classification accuracy, amongst the other combinations, which is at par with the ones from literature. However, this network also achieves significant reduction in (a) number of neurons (19 in this network cf. 63 or more elsewhere i.e. $> 3 \times$ improvement) and (b) synapses $16 \cdot 3 = 48$ in our network cf. $4 \cdot 10 + 10 \cdot 10 + 10 \cdot 3 = 170$ or more elsewhere, i.e $> 3.5 \times$ improvement) (c) number of training iterations (15 in our network cf. 450 or more elsewhere i.e. 30 ×). This can have major advantages in terms of power/energy and area required for operation in a hardware version. Since the timescales of training heavily depend on the size of the network, this network learns relatively quickly which can largely be attributed to its small size, for there are no hidden layers. In addition, the networks with other sensor combinations also achieve accuracies greater than 90 % as illustrated in TABLE I.

| Implementation | Input N1 | Hidden N2 | Output N3 | Total Synapses | Training Iterations | Accuracy |
|---|---|---|---|---|---|---|
| **Literature** | | | | | | |
| Spike Prop [4] | 50 | 10 | 3 | 530 | 1000 | 96.2 % |
| MATLAB BP [4] | 50 | 10 | 3 | 530 | 3750 | 95.8 % |
| Meta Neuron [4] | 4 | 10 x 10 | 3 | 170 | 1500 | 97.33 % |
| Quick Prop [4] | 50 | 10 | 3 | 530 | 450 | 97.33 % |
| **Our work** | | | | | | |
| 3 Linear Sensors | 12 | 0 | 3 | 36 | 15 | 94.7% |
| 4 Linear Sensors | 16 | 0 | 3 | 48 | 15 | 96.5% |
| 3 RBFs | 12 | 0 | 3 | 36 | 15 | 94% |
| 5 RBFs | 20 | 0 | 3 | 60 | 15 | 95.3% |
| 7 RBFs | 28 | 0 | 3 | 84 | 15 | 91.6% |
| **Improvement** | | 3 x | | 3.5 x | 30 x | Same |

TABLE I. Performance benchmark for the Fisher Iris classification dataset. Our work achieves significant reduction in the number of neurons, synapses and training iterations while performing at par with the state of the art. Note that the classification accuracy is stated as a ten-point average.

#### b) Wisconsin Cancer classification dataset

The *Wisconsin Cancer* dataset [19] consists of 699 samples with 9 features. The samples belong to two classes – *benign* and *malignant*. 20 samples each from *benign* and *malignant* classes were taken and used for training. The learned weights were tested on the entire dataset of 699 samples. Our network achieved an accuracy of 96.5% which is at par with that reported by [6].

| Implementation | Input N1 | Hidden N2 | Output N3 | Total Synapses | Training Iterations | Accuracy |
|---|---|---|---|---|---|---|
| **Literature** | | | | | | |
| Yan Xu et.al.[6] | 46 | 8 | 2 | 384 | 209 | 95.32% |
| **Our work** | | | | | | |
| 3 Linear Sensors | 27 | 0 | 2 | 54 | 15 | 96% |
| 4 Linear Sensors | 36 | 0 | 2 | 72 | 15 | 96.4% |
| 3 RBFs | 27 | 0 | 2 | 54 | 15 | 96.5% |
| 5 RBFs | 36 | 0 | 2 | 72 | 15 | 96.3% |
| 7 RBFs | 45 | 0 | 2 | 90 | 15 | 96.1% |
| **Improvement** | | 2 x | | 7 x | 14 x | Same |

TABLE II. Performance benchmark for the Wisconsin Breast Cancer classification problem. The network with 3 RBF sensors performs the best and achieves 2 × reduction in the number of neurons, 7 × reduction in the number of synapses and requires 14 × less training iterations as compared to [6].

#### c) Other datasets

We also tested our algorithm on two other datasets, whose results we will be briefly mentioning. First is the *wine dataset* [20] containing 178 samples belonging to three classes. Each sample possessing 13 features. A classification accuracy of 92% was achieved on this dataset by performing same experiment as in *Fisher Iris classification*. Second is the *heart-statlog dataset* [21] for predicting heart disease, consisting of 270 samples belonging to two classes. Each sample possessing 13 features (drawn from a larger set of 75 features). A classification accuracy of 79% was achieved after training a network with 7 RBF transforms in the first layer with 20 samples/class and higher threshold potentials for the second layer.

## B. Neuron vs. synapse interplay & co-design

Next, we present the interplay of neuron design (based on spiking threshold) and synaptic maximum current ($I_{max}$) to show that the optimal transformations of the initial data set (i.e. population coding) leads to maximum performance.

First, the LIF neuron possesses a current threshold (due to their leaky nature), i.e. only those feature values which translate to constant currents which lie above this threshold will be able to cause spiking in the input layer neurons. To obtain the value of the threshold current, we substitute $V$ with $V_t$ and $I$ with $I_{min}$ in Equation 7 and equate the LHS to zero, to obtain:

$$\frac{dV}{dt} = \frac{1}{C}(-g(V_t - E_l) + I_{th}) = 0$$

$$I_{th} = g(V_t - E_l) \quad (11)$$

$$I_{th} = 30\ nS \cdot (20 - (-70))mV = 2.7\ nA$$

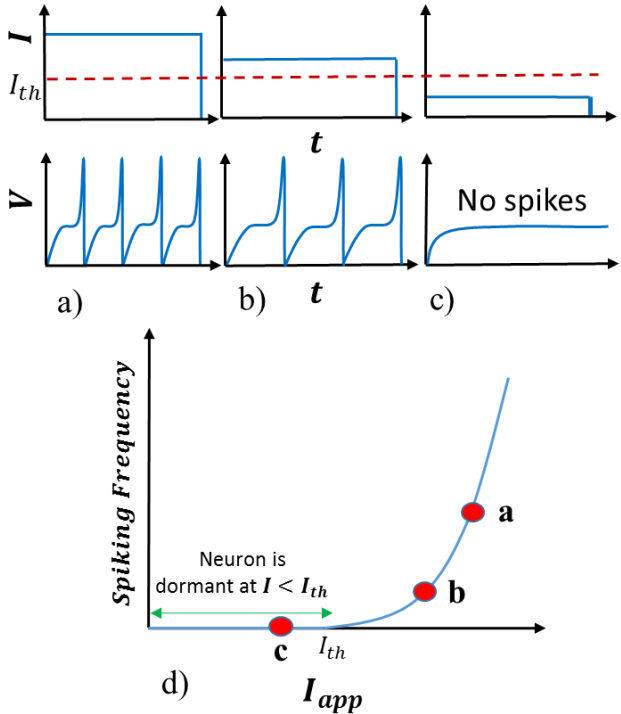

Fig. 5. The input output behaviour for a LIF neuron for different values of constant input currents. When the input current value is below the threshold (c), no spike is issued and the membrane potentials settles to constant steady value. d) Figure illustrating the non-linearity in the spiking behavior of the LIF neuron. For $I_{app} < I_{th}$, the neuron does not spike. This implies that it does not put any current down the line and completely blocks the incoming signal from propogating further.

For the values for the constants used in the previous section, $I_{th}$ comes out to be $2.7\ nA$. Once $I_{th}$ is fixed, we then sweep $I_{th}/I_{max}$ (the threshold value in the normalized relative scale) from 0.2 to 0.9 by varying $I_{max}$. The classification accuracy is found to be the maximum when the ratio is in the range 0.6~0.7 as shown in Fig. 6. We thus set the value of $I_{max}$ to be $4\ nA$ for which, the ratio comes out to be in the above range.

$$\frac{I_{th}}{I_{max}} = \frac{2.7\ nA}{4\ nA} = 0.675$$

Next, we try to find the reason for such sharp sensitivity of the classification accuracy on the above ratio by exploring its effect on the linear sensors described earlier. Since any feature value less than 0.675 (on the normalized scale) would *not excite any response* in the input layer, this has a threshold effect on the sensors, as illustrated in Fig. 5. This behavior thus plays a vital role in sensor design, as now, for the choice of $\frac{I_{th}}{I_{max}} = 0.675$, the first three transformations turn out to be perfectly mutually exhaustive (as they leave a negligible gap on the x-axis) and exclusive (they do not overlap), as illustrated in Fig. 7(e).

The former is essential as all the sensors combined should be able to pick data lying at all parts of the input spectrum whereas the latter ensures that the sensors are not intermixed and hence redundant. As $I_{th}/I_{max}$ increases beyond 0.7, the unattended gap $\Delta x$ increases leading to data-loss and hence a decline in performance, as can be seen Fig. 6. Similarly, when $I_{th}/I_{max}$ goes below 0.6, the sensors start to overlap leading to intermixing and an eventual drop in performance.

This analysis also explains the ineffectiveness of using just the raw normalized data (which is equivalent to using only the first transformation) as it only "*senses*" the high data values which results in a complete "*loss*" of the low and middle parts of the spectrum as is clearly seen in Fig. 7(a). In addition, it explains why the fourth transformation is largely redundant and not very necessary as it covers a part of the range already covered by transformations one and two together.

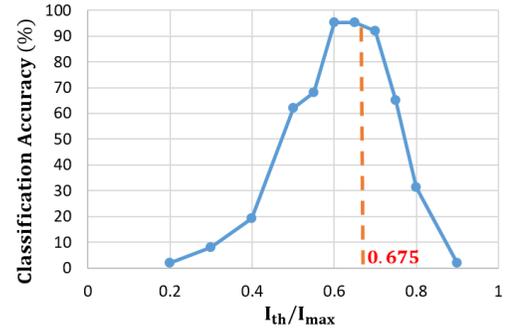

Fig. 6. The varation of classification accuracy with the ratio $I_{th}/I_{max}$. The peak is observed for the range of values for which the sensors are mutuallye exclusive and mutually exhaustive. This serves as a design curve for setting the value of $I_{max}$ for best performance.

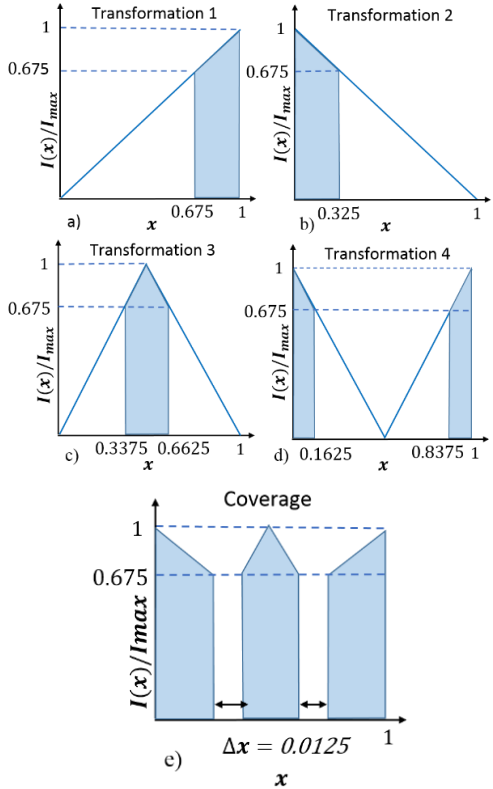

Fig. 7. The four linear transformations after applying the threshold filter. (e) shows the effective portions of the sensors in (a), (b) and (c), which mutually and exhaustively cover almost the entire input range (negligible gaps in between), leaving the sensor in (d) redundant.

Thus, the first three transformations are necessary and sufficient for good performance. Excluding any of them makes the performance drop significantly, as they help cover the entire spectrum of input values, whereas the presence of the fourth transform has a very minor effect on performance, as can be seen from TABLES I & II. Fig. 6 serves as a design curve to set the value of $I_{max}$ for best performance. Thus, the efficient network is based on important population coding and synapse-neuron co-design.

## IV. FAST RECOGNITION EVALUATION AND ROBUSTNESS

### A. Current Space Analysis

Assessment of the extent of learning in SNNs is usually done using the standard recognition tests which involves simulating the actual neuronal dynamics. However, this methodology has several drawbacks. First, it is computationally intensive. For instance, performing one round of testing for all the 699 samples in the *Wisconsin Breast Cancer* classification problem takes about $150s$ in MATLAB. A faster evaluation is hence attractive. Second, it only gives a single output, namely the classification accuracy, which does not indicate robustness of learning. In this section, we develop simple and efficient method to evaluate the performance of the network without standard recognition tests. Finally, we also develop a learning robust-ness metric for noise tolerant recognition performance.

*a) Methodology*

Let us take the case when our network has $M$ input neurons and $N$ output neurons. Each of the $N$ output neurons corresponds to one particular class. For each sample, let $I_1, I_2, ..., I_M$ be the $M$ current values being fed to the $M$ input layer neurons. Let us consider the $j^{th}$ output neuron, corresponding to the $j^{th}$ class. This neuron is connected to each of the $M$ input neurons. Let $w_{i,j}$ be the weight acquired after learning, by the synapse connecting the $i^{th}$ input neuron to the $j^{th}$ output neuron. Therefore, for this particular sample, the aggregate total current received by the $j^{th}$ output neuron can be approximated as:

$$I_{total,j} \equiv \sum_{i=1}^{M} w_{i,j} \cdot I_i \cdot u(I_i - I_{th}) \quad (12)$$

where $u$ is the unit step function to account for the threshold behavior of the LIF neurons, as explained in the previous section. Current values lower than $I_{th}$ do not cause even a single spike in the first layer neurons and thus do not contribute to the current between the first and the second layers. Thus, the SNN creates an inherent mapping of the sample points into a $N$ dimensional space, with the magnitude along each dimension being equal to the current supplied to the corresponding neuron. For instance, if $N$ equals 2, the coordinates for the sample in the *current space* will be given by

$$\boldsymbol{I} = I_{total,1}\,\hat{x} + I_{total,2}\,\hat{y} \quad (13)$$

We next describe two performance metrics which can be derived from this space, in the following sections.

*b) Merit Figure*

We argue that if the classification were to proceed correctly, the samples actually belonging to class $j$ should feed a large current to the $j^{th}$ neuron and little current to the other output neurons. It is only then that the probability of the $j^{th}$ neuron to spike and for the others not to spike during the observed testing interval is enhanced. On the other hand, if the sample happens to feed similar amounts of currents to two or more output neurons, then each of them are equally likely to spike and the sample is likely to be misclassified. This can then be extended for each of the output neurons.

For the case when $N$ equals two and if the above sample happens to belong to the first class (say), the following inequality should be satisfied for correct classification to occur.

$$I_{total,1} > I_{total,2}$$

In other words, since the sample belongs to the first class, it should feed a larger current to the first neuron than to the second neuron. Thus, each sample should ideally fall into a particular subspace corresponding to its class so that the above inequality is satisfied. Hence, we define the *merit figure* as the proportion of the samples, in percentage, which lie in the correct subspace.

$$Merit\ Figure \equiv \left(\frac{\#\ of\ samples\ in\ correct\ subspace}{Total\ \#\ of\ samples}\right) \cdot 100 \quad (14)$$

*c) Illustration for the Wisconsin Breast Cancer classification problem*

We use the nine RBF sensors for the analysis of the *Wisconsin Cancer dataset* in this section. The samples from this dataset belong to two classes – *benign* and *malignant*. Thus, $N = 2$ here and the *current space* is of dimension two. Fig. 8 illustrates such a space with the 45° line partitioning the entire space into two subspaces. The *benign* samples are shown as *red* crosses (with their mean as a *big red* dot) while the *malignant* samples are shown as *blue* circle (with their mean as a *big blue* dot). Fig. 8 (a) shows the current space prior to any training, with the weights equal to their initialized value. The samples are randomly interwoven around the 45° line and are virtually indistinguishable. Fig. 8 (b) illustrates the scenario after one training epoch. The *merit figure* improves significantly. We continue the training till 14 epochs, after which we perform a round of testing. The classification accuracy is found to be within 1% of the *merit figure* obtained from our analysis. Thus, the *merit figure* shows a high correlation with the actual classification accuracy (to be discussed further). A concise visualization of performance improvement during learning is achieved.

*d) Robustness Metric*

Although the *merit figure* serves as an accurate indicator for network performance when the system is ideal, it does not assess the robustness against noise introduced in the weights. After all, the synaptic weights are modelled as RRAMs which are prone to minor variations. We observe that ideally, the samples should lie well within their respective sub-spaces as then, even if they move around a little, they would not flip over to the wrong subspace. However, if they lie close to the boundary, they are likely to end up in the wrong subspace leading to a drop in performance. We thus define the *robustness metric* as

$$Robustness\ Metric \equiv \min(\boldsymbol{d}) \cdot 100 \quad (15)$$

where $\boldsymbol{d}$ is the normalized distance of the means of samples belonging from each class, from their corresponding sub-space boundary, where the normalization is done with the maximum possible separation that the means could have. Thus, the closer this metric is to hundred, the farther the samples are from the boundary and hence the more robust the system is against minor variations in the weights. However, if this metric is close to zero or even negative, it implies that the samples are either very close to the boundary or even in the wrong side of it, implying that the system is very sensitive to variations in weights. In the ideal scenario, we would like the means to lie at the extremes of the axes, as that would mean pushing a high current to the correct neuron and negligible current to the other neurons.

Fig. 8 (a) shows that $d_{blue}$ is negative as the blue mean lies in the wrong subspace while $d_{red}$ is positive but very close to zero. Fig. 8 (b) illustrates the scenario after one training epoch. The *merit figure* improves significantly while $d_{blue}$ flips signs. A key observation here is that although 93% percent of samples move to their correct subspaces with only one epoch of training, the subsequent training epochs make the learning more robust as the two means move further into their subspaces as can be seen in Fig. 8 (c). $d_{blue}$ increases further suggesting a more thorough learning taking place, thereby making the system more robust. The analysis thus illustrates that if the system is ideal, one training epoch is sufficient to achieve high classification accuracies. However, if the system is non-ideal and prone to minor variations in the weights, the training should be repeated for more epochs to make the system more fault-tolerant. The *robustness metric* gives a quantitative measure of such a tolerance. Thus, between one and fourteen training epochs, the *merit figure* improves only slightly ( 93 % → 96 %) while the robustness metric improves significantly (7.5 → 17.92). To evaluate the impact of the robustness, we add increasing noise to study its effect of performance.

We refer to the systems in Fig. 8 (b) and (c) as system *A* and system *B* respectively. Further, we introduce a zero mean Gaussian noise ($\sigma = 300$) in the learnt weights of the two systems with different extents of training to verify our claims. The physical origin of this noise could be variations in memristor behavior over time [22]. Fig. 8 (d) and (e) show the original means as well as the ten scattered means for system *A* and system *B* respectively.

*e) Correlation, Efficiency and Noise Tolerance Measure*

First, based on large number of recognition events, we observe that the *merit figure* and the classification accuracy have a high correlation coefficient of 0.98, thereby empirically justifying that the *merit figure* is indeed a good approximate of the actual classification accuracy, as seen in Fig. 9(a).

A significant reduction (15000 ×) in its computation time is achieved viz. the standard recognition test, as shown in Fig. 9 (b). For instance, the computation of the *merit figure* for all samples of the *Wisconsin Breast Cancer* data-set takes a mere 0.01s in MATLAB as it essentially is the equivalent of DC analysis as opposed to transients in standard recognition which would take 150 s.

Next, we use the *merit figure* to gauge the performance degradations of the two systems with increasing noise variance. For this, we sweep $\sigma$ from 10 to 300 and at each step, and report a 1000 point average for the *merit figure*, to account for the stochasticity in the noise. We observe that the system after fourteen training epochs (system *B*) is indeed more fault-tolerant, as can be clearly seen in Fig. 9 (b). For any value of $\sigma$, system *B* reports a higher *merit figure* than system *A*.

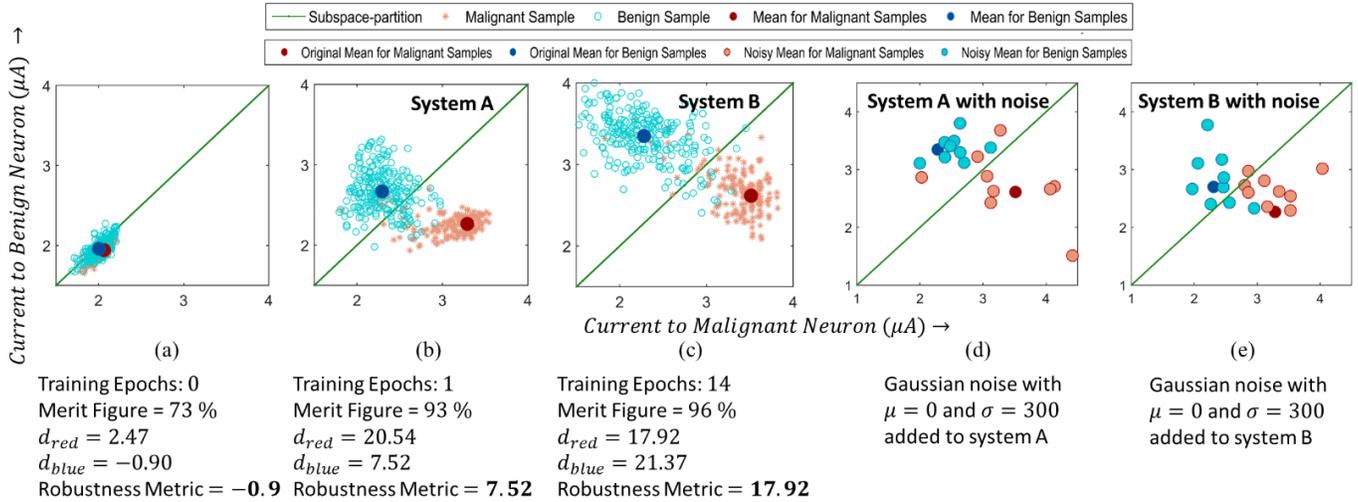

Fig. 8. The Wisconsin Cancer data samples in the current space along with their means illustrating the evolution of the *merit figure* and the Separation Metric ($d$). (a) The initial configuration prior to training with a low *merit figure* and negative value for $d_{blue}$ (b) After one training epoch, the *merit figure* rises while the two means fall into their respective sub-spaces (c) After 14 training epochs, the *merit figure* has a minor increment while $d_{blue}$ increases further. The classification accuracy at this stage closely resembles the *merit figure*. (d) and (e) The scattered means for system $A$ and system $B$ respectively after adding a zero mean 300 variance noise to the learnt weights.

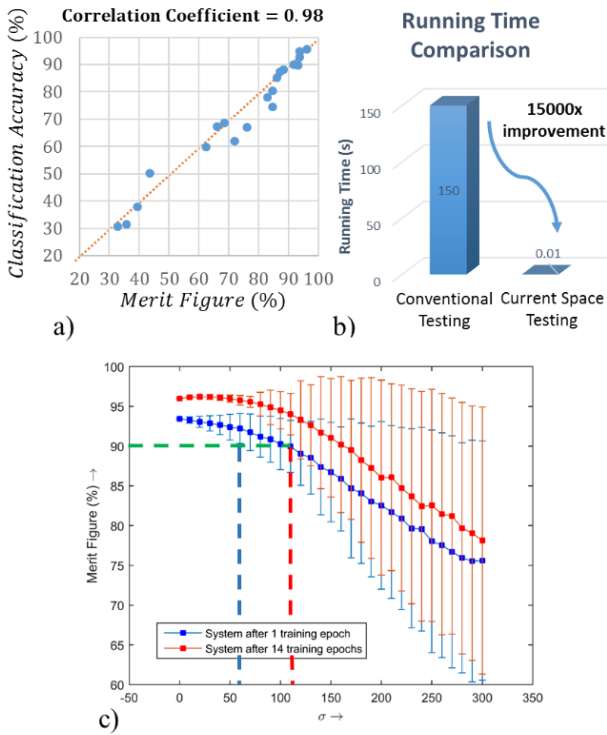

Fig. 9. (a) A high correlation coefficient of 0.98 is observed between classification accuracy and merit figure for the 20 random data points, 10 each from systems $A$ and $B$ (b) A comparison of computational times of the two testing methodologies (c) Performance degradation with noise introduced in the learned weights of the two systems with different extents of training. System $B$ performs better, as predicted by the *robusteness metric*. Note that the *merit figure* is reported as a 1000 point average to account for the stochasticity in the noise. The vertical lines show the variance of the merit figures for each value of $\sigma$.

At a threshold of 90% accuracy, the noise tolerance is $\sigma =50$ for case A compared to $\sigma =100$ for case B (2x change – which is proportional to the *robustness metric* change of 2.5×). In comparison, the initial performance only shows a 3% difference (i.e. very weak dependence).

V. CONCLUSION

Towards the goal of developing a simple SNN for classification tasks by STDP based SNN, which can be implemented on cross-bar array devices- we demonstrate some key enablers. First, we have achieved significant reduction in complexity of network structure (>3.5× reduction in neurons & synapse, and 30 × reduction in epochs) without compromising on performance. The efficient network is based on effective population coding and synapse-neuron co-design. Second, an efficient and detailed method to evaluate SNN performance is proposed and validated. The output currents space provides a detailed visualization of the time-evolution of learning in SNNs. The method provides a 15000 × computational efficiency as well as excellent accuracy (correlation of 0.98 to conventional testing). It also accelerates stochastic simulations. Third, a robust-ness metric is developed. For comparable performance (3% difference), robustness metric shows a 2.5× difference which is similar in order of magnitude to the difference in noise tolerance (2×). Thus, we show that the learning is more robust (that is, noise tolerance performance improves) with epochs.

## VI. REFERENCES

[1] G. J. Siegel, B. W. Agranoff, and R. W. Albers, Eds., *Basic Neurochemistry: Molecular, Cellular and Medical Aspects*. Philadelphia, PA: Lippincott-Raven, 1999.

[2] Y. Xu, X. Zeng, L. Han and J. Yang, "A supervised multi-spike learning algorithm based on gradient descent for spiking neural networks," *Neural Networks,* pp. 99-113, 2013.

[3] S. Ghosh-Dastidar and H. Adeli, "A new supervised learning algorithm for multiple spiking neural networks with application in epilepsy and seizure detection," *Neural Networks,* p. 1419–1431, 2009.

[4] J. Xin and M. Embrechts, "Supervised learning with spiking neural networks," in *IJCNN*, 2001.

[5] I. Sporea and A. Grüning, "Supervised Learning in Multilayer Spiking Neural Networks," *Neural Compuation,* pp. 473-509, 2012.

[6] Y. Xu, X. Zeng and S. Zhong, "A New Supervised Learning Algorithm for Spiking Neurons," *Neural Computation,* p. 1472–1511, 2013.

[7] A. Kasinski and P. Ponulak, "Comparison of Supervised Learning Methods for Spike Time Coding in Spiking Neural Networks," *Int. J. Appl. Math. Comput. Sci.,* pp. 101-113, 2006.

[8] P. Ponulak, "ReSuMe - New Supervised Learning Method," Poznan, 2005.

[9] B. Ruf and M. Schmitt, "Learning Temporally Encoded Patterns in Networks," *Neural Processing Letters,* pp. 9-18, 1997.

[10] A. Shukla, V. Kumar, U. Ganguly - A Software-equivalent Hardware approach to Spiking Neural Network based Real-time Learning using RRAM array demonstration in SPICE* - submitted for IJCNN 2017

[11] S.M. Bohte, J.N. Kok, and H. La Poutre (2002) - Error Backpropagation in Temporally Encoded Networks of Spiking Neurons

[12] M.N. Shadlen and W.T. Newsome - The variable discharge of cortical neurons: implications for connectivity, computation and information coding, *Neuronsci,* 18(1998) 3870-389

[13] K.C. Zhang, **I.**Ginzburg, B.LI McNaughton - Interpreting Neuronal Population Activity by Reconstruction: Unified Framework With Application to Hippocampal Place Cells, *J. Neurophysiol.* 79: 1017–1044, 1998

[14] A. Gupta and L. N. Long, "Character Recognition using Spiking Neural," in *IEEE Neural Networks Conference*, Orlando, FL, 2007.

[15] Christof Koch, Idan Segev, (1999) - Methods in neuronal modeling; from ions to networks (2nd ed.). Cambridge, Massachusetts: MIT Press. p. 687. ISBN 0-262-11231-0

[16] R. Gütig and S. Haim, "The tempotron: a neuron that learns spike timing-based decisions," *Nature Neuroscience,* pp. 420-428, 2006.

[17] Guetig R., Aharonov R., Rotter S. and Sompolinsky H (2003) - Learning input correlations through non-linear temporally asymmetric Hebbian plasticity

[18] Newman, D.J., Hettich, S., Blake, C.L., & Merz, C.J. (1998). UCI repository of machine learning databases. Irvine, CA: Department of Information and Computer Science, University of California. Available: http://www.ics.uci.edu/mlearn/MLRepository.html.

[19] O. L. Mangasarian, W. N. Street and W. H. Wolberg, "Breast Cancer Diagnosis and Prognosis via Linear Programming," AAAI Technical Report SS-94-01, 1994.

[20] Forina, M. et al, PARVUS - An Extendible Package for Data Exploration, Classification and Correlation. Institute of Pharmaceutical and Food Analysis and Technologies, Via Brigata Salerno, 16147 Genoa, Italy

[21] Heart-statlog data - source: V.A. Medical Center, Long Beach and Cleveland Clinic Foundation: Robert Detrano

[22] Boubacar Traoré, Philippe Blaise, Elisa Vianello, Helen Grampeix, Simon Jeannot, Luca Perniola, Barbara De Salvo, and Yoshio Nishi - On the Origin of Low-Resistance State Retention Failure in HfO2-Based RRAM and Impact of Doping/Alloying - IEEE TRANSACTIONS ON ELECTRON DEVICES, 2015